\documentclass{article}

\usepackage{microtype}
\usepackage{graphicx}
\usepackage{subfigure}
\usepackage{booktabs} 
\usepackage{color}

\usepackage{amsthm,amsmath,amssymb}
\usepackage{changepage}

\newcommand\transp{^\intercal\kern-\scriptspace}

\usepackage{hyperref}

{\vspace{-\topsep}\begin{itemize}\itemsep1pt \parskip0pt \parsep0pt \topsep1pt}
{\end{itemize}\vspace{-\topsep}}

\usepackage[accepted]{icml2018}

\icmltitlerunning{Retrieval-augmented Convolutional Neural Networks against adversarial examples}

\begin{document}

\twocolumn[
\icmltitle{Retrieval-Augmented Convolutional Neural Networks \\ for Improved Robustness against Adversarial Examples}



\icmlsetsymbol{equal}{*}

\begin{icmlauthorlist}
\icmlauthor{Jake (Junbo) Zhao}{nyu,fair}
\icmlauthor{Kyunghyun Cho}{nyu,fair,cifar}
\end{icmlauthorlist}

\icmlaffiliation{nyu}{New York University}
\icmlaffiliation{fair}{Facebook AI Research}
\icmlaffiliation{cifar}{CIFAR Azrieli Global Scholar}

\icmlcorrespondingauthor{Jake (Junbo) Zhao}{j.zhao@nyu.edu}

\icmlkeywords{Machine Learning, ICML}

\vskip 0.3in
]



\printAffiliationsAndNotice{}  

\begin{abstract}
We propose a retrieval-augmented convolutional network and propose to train it with local mixup, a novel variant of the recently proposed mixup algorithm. The proposed hybrid architecture combining a convolutional network and an off-the-shelf retrieval engine was designed to mitigate the adverse effect of off-manifold adversarial examples, while the proposed local mixup addresses on-manifold ones by explicitly encouraging the classifier to locally behave linearly on the data manifold. Our evaluation of the proposed approach against five readily-available adversarial attacks on three datasets--CIFAR-10, SVHN and ImageNet-- demonstrate the improved robustness compared to the vanilla convolutional network. 
\end{abstract}

\section{Introduction}

Since the initial investigation by \citet{szegedy2013intriguing}, adversarial examples have drawn a large interest. Various methods for both generating adversarial examples as well as protecting a classifier from them have been proposed (see Sec.~\ref{sec:attack}--\ref{sec:related} for more details.) Adversarial examples exist due to misbehaviors of a classifier in some regions of the input space and are generated often by finding a point in such a region using optimization.

According to \citep{gilmer2018adversarial}, adversarial examples can be categorized into those off the data manifold, which is defined as a manifold on which training examples lie, and those on the data manifold. Off-manifold adversarial examples occur as the classifier does not have a chance to observe any off-manifold examples during training, which is a natural consequence from the very definition of the data manifold. On-manifold adversarial examples however exist between training examples on the data manifold. There are two causes behind this phenomenon; (1) the sparsity of training examples and (2) the non-smooth behavior of the classifier on the data manifold.

In this paper, we propose to tackle both off- and on-manifold adversarial examples by incorporating an off-the-shelf retrieval mechanism which indexes a large set of examples and training this combination of a deep neural network classifier and the retrieval engine to behave linearly on the data manifold using a novel variant of the recently proposed mixup algorithm~\citep{zhang2017mixup}, to which we refer as ``local mixup.''

The retrieval mechanism efficiently selects a subset of neighboring examples from a candidate set near the input. These neighboring examples are used as a local approximation to the data manifold in the form of a feature-space convex hull onto which the input is projected. The classifier then makes a decision based on this projected input. This addresses off-manifold adversarial examples. Within this feature-space convex hull, we encourage the classifier to behave linearly by using local mixup to further address on-manifold adversarial examples.

We evaluate the proposed approach, called a retrieval-augmented classifier, with a deep convolutional network~\citep{lecun1998gradient} on object recognition. We extensively test the retrieval-augmented convolutional network (RaCNN) on datasets with varying scales; CIFAR-10~\citep{krizhevsky2009learning}, SVHN~\citep{netzer2011reading} as well as ImageNet~\citep{deng2009imagenet}, against five readily-available adversarial attacks including both white-box (FGSM, iFGSM, DeepFool and L-BFGS) and black-box attacks (Boundary). Our experiments reveal that the RaCNN is more robust to these five attacks than the vanilla convolutional network.

\section{Retrieval-Augmented CNN}
\label{sec:model}

\citet{gilmer2018adversarial} have recently demonstrated that adversarial examples exist both on and off the data manifold in a carefully controlled setting in which examples from two classes are placed on two disjoint spheres. This result suggests that it is necessary to tackle both types of adversarial examples to improve the robustness of a deep neural network based classifier to adversarial examples. In this section, we describe our approach toward building a more robust classifier by combining an off-the-shelf retrieval engine and a variant of the recently proposed mix-up learning strategy.

\subsection{Setup}

Let $D'=\left\{ (x'_1,y'_1), \ldots, (x'_M,y'_M) \right\}$ be a candidate set of examples. This set may be created as a subset from a training set $D=\left\{ (x_1, y_1), \ldots, (x_N, y_N) \right\}$ or may be an entire separate set. We use $D'$ as a proxy to the underlying data manifold. 

$k(x, x')$ is a distance function that measures the dissimilarity between two inputs $x$ and $x'$. In order to facilitate the use of an off-the-shelf retrieval engine, we use 
\begin{align}
\label{eq:distance}
k(x, x')=\|\phi'(x) - \phi'(x')\|^2,
\end{align} 
where $\phi'$ is a predefined, or pretrained, feature extractor. We assume the existence of a readily-available retrieval engine $F$ that takes $x$ as input and returns the $K$ nearest neighbors in $D'$ according to $k(x, x')$. 

We then have a deep neural network classifier composed of a feature extraction $\phi$ and a classifier $g$. This classifier is trained on a training set $D$, taking into account the extra set $D'$ and the retrieval engine.

\subsection{Inference}

In this setup, we first describe the forward evaluation of the proposed network. This forward pass is designed to handle adversarial examples ``off'' the data manifold by projecting them onto the data manifold.

\paragraph{Local Characterization of Data Manifold}

Given a new input $x$, we use the retrieval engine $F$ to retrieve the examples $x'_k$'s from $D'$ that are closest to $x$: $F(x)=\left\{ x'_1, \ldots, x'_K \right\}$. We then build a feature-space convex hull by
\[
\mathcal{C}(F(x)) = \left\{ \sum_{k=1}^K \alpha_k \phi(x'_k) \left| \sum_{k=1}^K \alpha_k = 1 \wedge \forall k: \alpha_k \geq 0 \right.\right\}.
\]
As observed earlier, linear interpolation of two input vectors in the feature space of a deep neural network often corresponds to a plausible input vector, unlike when interpolation was done in the raw input space~\citep[see, e.g.,][]{bengio2013better,kingma2013auto,radford2015unsupervised}. Based on this observation, we consider the feature-space convex hull $\mathcal{C}(F(x))$ as a reasonable local approximation to the underlying data manifold.  

\paragraph{Trainable Projection}

Exact projection of the input $x$ onto this convex hull $\mathcal{C}(F(x))$ requires expensive optimization,
especially in the high-dimensional space. As we consider a deep neural network classifier, the dimension of the feature space $\phi'$ could be hundreds or more, making this exact projection computationally infeasible. Instead, we propose to learn a goal-driven projection procedure based on the attention mechanism~\citep{bahdanau2014neural}. 

We compare each input $x'_k \in F(x)$ against $x$ and compute a score:
\[
\beta_k = \phi(x'_k)^\top U \phi(x),
\]
where $U$ is a trainable weight matrix~\citep{luong2015effective}. These scores are then normalized to form a set of coefficients:
$\alpha_k = \frac{\exp(\beta_k)}{\sum_{k'=1}^K \exp(\beta_{k'})}.$
These coefficients $\alpha_k$'s are then used to form a projection point of $x$ in the feature-space convex hull $\mathcal{C}(F(x))$:
\begin{align*}
\mathcal{P}(x) = \mathcal{P}_{\mathcal{C}(F(x))}(x) = \sum_{k=1}^K \alpha_k \phi(x'_k).
\end{align*}
This trainable projection could be thought of as learning to project an off-manifold example on the locally-approximated manifold to maximize the classification accuracy.

\paragraph{Classification}

The projected feature $\mathcal{P}_{\mathcal{C}(F(x))}(x)$ now represents the original input $x$ and is fed to a final classifier $g$. In other words, we constrain the final classifier to work only with a point inside a feature-space convex hull of neighboring training examples. This constraint alleviates the issue of the classifier's misbehavior in the region outside the data manifold up to a certain degree.\footnote{
The quality of the local approximation may not be uniformly high across the input space, and we do not claim that it solves the problem of off-manifold adversarial examples.
}

\subsection{Training}

The output of the classifier $g(\mathcal{P}(x))$ is almost fully differentiable with respect to the classifier $g$, both of the features extractors ($\phi'$ and $\phi$) and the attention weight matrix $U$, except for the retrieval engine $F$.\footnote{
We believe the introduction of this non-differentiable, black-box retrieval engine further contributes to the increased robustness against white-box attacks.
} 
This allows us to train the entire pipeline in the previous section using backpropagation~\citep{rumelhart1986learning} and gradient-based optimization. 

\paragraph{Local Mixup}

This is however not enough to ensure the robustness of the proposed approach to on-manifold adversarial examples. During training, the classifier $g$ only observes a very small subset of any feature-space convex hull. Especially in a high-dimensional space, this greatly increase the chance of the classifier's misbehavior within these feature-space convex hulls, as also noted by \citet{gilmer2018adversarial}. In order to address this issue, we propose to augment learning with a local variant of the recently proposed mix-up algorithm~\citep{zhang2017mixup}. 

The goal of original mixup is to encourage a classifier to act linearly between any pair of training examples. This is done by linearly mixing in two randomly-drawn training examples and creating a new linearly-interpolated example pair during training. Let two randomly-drawn pairs be $(x_i, y_i)$ and $(x_j, y_j)$, where $y_i$ and $y_j$ are one-hot vectors in the case of classification. Mixup creates a new pair $(\lambda x_i + (1-\lambda) x_j, \lambda y_i + (1-\lambda) y_j)$ and uses it as a training example, where $\lambda \in [0,1]$ is a random sample from a beta distribution. We call this original version {\it global} mixup, as it increases the linearity of the classifier between any pair of training examples. 

It is however unnecessary for our purpose to use global mixup, as our goal is to make the classifier better behave (i.e., linearly behave) within a feature-space convex hull $\mathcal{C}(F(x))$. Thus, we use a {\it local} mixup in which we uniformly sample the convex coefficients $\alpha_k$'s at random to create a new mixed example pair $(\sum_{k=1}^K \alpha_k \phi(x'_k), \sum_{k=1}^K \alpha_k y'_k)$. We use the Kraemer Algorithm~\citep[see Sec. 4.2 in][]{smith2004sampling}. 

\paragraph{Overall}

We use stochastic gradient descent (SGD) to train the proposed network. At each update, we perform $N_{\text{CE}}$ descent steps for the usual classification loss, and $N_{\text{MU}}$ descent steps for the proposed local mixup. 

\subsection{Retrieval Engine $F$}

The proposed approach does not depend on the specifics of a retrieval engine $F$. Any off-the-shelf retrieval engine that supports dense vector lookup could be used, enabling the use of a very large-scale $D'$ with latest fast dense vector lookup algorithms, such as FAISS~\citep{johnson2017billion}. In this work, we used a more rudimentary retrieval engine based on locality-sensitive hashing~\citep[LSH; see, e.g.,][]{datar2004locality} with a reduced feature dimension using random projection~\citep[see, e.g.,][and references therein]{bingham2001random}, as the sizes of candidate sets $D'$ in the experiments contain approximately 1M or less examples. The key $\phi'(x)$ from Eq.~\eqref{eq:distance} was chosen to be a pretrained deep neural network without the final fully-connected classifier layers \citep{krizhevsky2012imagenet,he2016deep}.

\section{Adversarial Attack}
\label{sec:attack}

\subsection{Attack Scenarios}

\paragraph{Scenario 1 (Direct Attack)} In this work, we consider the candidate set $D'$ and the retrieval engine which indexes it to be ``hidden'' from the outside world. This property makes a usual white-box attack more of a {\it gray-box} attack in which the attacker has access to the entire system except for the retrieval part. This is our first attack scenario.

\paragraph{Scenario 2 (Retrieval Attack)} Despite the hidden nature of the retrieval engine and the candidate set, it is possible for the attacker to confuse the retrieval engine, if she/he could access the feature extractor $\phi'$. We furthermore give the attacker the access not only to $\phi'$ but the original classifier $g'$ which was tuned together with $\phi'$. This allows the attacker to create an adversarial example on $g'(\phi'(x))$ that could potentially disrupt the retrieval process, thereby fooling the proposed network. Although this is unlikely in practice, we test this second scenario to investigate the possibility of compromising the retrieval engine.

\subsection{Attack Methods}

Under each of these scenarios, we evaluate the robustness of the proposed approach on the five widely used/tested adversarial attack algorithms including both white-box and black-box attacks. They are fast gradient sign method~\citep[FGSM,][]{goodfellow2014explaining}, its iterative variant~\citep[iFGSM,][]{kurakin2016adversarial}, DeepFool~\citep{moosavi2016deepfool}, L-BFGS~\citep{tabacof2016exploring} and Boundary~\citep{brendel2017decision}. We acknowledge that this is not an exhaustive list of attacks, however find it to be extensive enough to empirically evaluate the robustness of the proposed approach.

\paragraph{Fast Gradient Sign Method (FGSM)}

FGSM creates an adversarial example by adding the scaled sign of the gradient of the loss function $L$ computed using a target class $\hat{y}$ to the input:
\[
    x' = x + \epsilon \cdot \text{sign}(\nabla_x L(x, \hat{y})),
\]
where the scale $\epsilon$ controls the difference between the original input $x$ and its adversarial version $x'$. This is a white-box attack, requiring the availability of the gradient of the loss function with respect to the input. 

\paragraph{Iterative FGSM (iFGSM)}

iFGSM improves upon the FGSM by iteratively modifying the original input $x$ for a fixed number $S$ of steps. At each step,
\[
x^{(s)} = x^{(s-1)} + \frac{\epsilon}{S} \text{sign}(\nabla_x L(x^{(s-1)}, y)),
\]
where $s=1,\ldots, S$ and $x^0=x$. Similarly to the FGSM, the iFGSM is a white-box attack.

\paragraph{DeepFool} 

\citet{moosavi2016deepfool} proposed to create an adversarial example by finding a residual vector $r \in \mathbb{R}^{\text{dim}(x)}$ with the minimum $L_p$-norm with the constraint that the output of a classifier must flip. They presented an efficient iterative procedure to find such a residual vector. Similarly to the FGSM and iFGSM, this approach relies on the gradient of the classifier's output with respect to the input, and is hence a white-box attack.

\paragraph{L-BFGS}

\citet{tabacof2016exploring} proposed an optimization-based approach, similar to DeepFool above, however, more explicitly constraining the input to lie inside a tight box defined by training examples. They use L-BFGS-B~\citep{zhu1997algorithm} to solve this box-constrained optimization problem. This is also a white-box attack.

\paragraph{Boundary}

\citet{brendel2017decision} proposed a powerful black-box attack, or more specifically decision-based attack, that requires neither the gradient of a classifier nor the predictive distribution. It only requires the final decision of the classifier. Starting from an adversarial example, potentially far away from the original input, it iteratively searches for a next adversarial example that has a smaller difference to the original input. This procedure guarantees the reduction in the difference by rejecting any step that neither decreases the difference nor makes the example not adversarial.

\paragraph{Implementation}

We use Foolbox\footnote{
Available at \url{http://foolbox.readthedocs.io/en/latest/}. Revision 2d468cb6.
} 
released by \citet{rauber2017foolbox}. Whenever necessary for further analysis, such as the accuracy per the amount of adversarial perturbation, we implement some of these attacks ourselves.

\section{Related Work}
\label{sec:related}

Since the phenomenon of adversarial examples was noticed by \citet{szegedy2013intriguing}, there have been a stream of attempts at making a deep neural network more robust. Most of the existing work are orthogonal to the proposed approach here and could be used together. We however detail them here to demonstrate similarities and contrasts against our approach.

\subsection{Input Transformation}

An off-manifold adversarial example can be avoided, if it could be projected onto the data manifold, characterized by training examples. This could be thought of as transforming an input. There have been two families of algorithms in this direction. 

\paragraph{Data-Independent Transformation}

The first family of defense mechanisms aims at reducing the input space so as to minimize regions that are off the data manifold. \citet{dziugaite2016study} demonstrated that JPEG-compressed images suffer less from adversarial attacks. \citet{lu2017no} suggest that trying various scaling of an image size could overcome adversarial attacks, as they seem to be sensitive to the scaling of objects. \citet{guo2017countering} uses an idea of compressed sensing to transform an input image by reconstructing it from its lower-resolution version while minimizing the total variation~\citep{rudin1992nonlinear}. More recently, \citet{buckman2018thermometer} proposed to discretize each input dimension using thermometer coding. These approaches are attractive due to their simplicity, but there have some work showing that it is often not enough to defend against sophisticated adversarial examples~\citep[see, e.g.,][]{shin2017jpeg}.

\paragraph{Data-Dependent Transformation}

On the other hand, various groups have tried using a data-dependent transformation mostly relying on density estimation. \citet{gu2014towards} used a denoising autoencoder~\citep{vincent2010stacked} to push an input back toward the data manifold. \citet{samangouei2018defensegan} and \citet{song2017pixeldefend} respectively use a pixelCNN~\citep{van2016conditional} and generative adversarial network~\citep{goodfellow2014generative} to replace an input image with a nearby, likely image. Instead of using a separately trained generative model, \citet{guo2017countering} uses a technique of image quilting~\citep{efros2001image}. These approaches are similar to our use of a retrieval engine over the candidate set. They however do not attempt at addressing the issue of misbehaviors of a classifier on the data manifold.

\begin{table*}[h!]
\caption{The CIFAR-10 classifiers' robustness to the adversarial attacks in the Scenario 2 (Retrieval Attack)
\label{tab:cifar10}}
\begin{center}
\small
\begin{tabular}{c|c|ccc|ccc|ccc} 
\toprule
 & Clean & & FGSM & & &  iFGSM & & & DeepFool \\
\midrule
$\overline{L_2}$ & 0 & 1e-04 & 2e-04 & 4e-04 & 1e-05 & 2e-05 & 8e-05 & 1e-05 & 2e-05  & 8e-05  \\ 
\midrule
Baseline & \textbf{85.15} & 14.05 & 7.5 & 4.22 & 55.2 & 26.17 & 2.59 & 26.04 & 11.72 & 0.34 \\ 
\midrule
RaCNN-K5 & 72.57 & 42.97 & 34.29 & 24.55 & 72.57 & 72.48 & 45.46 & 64.34 & 61.34 & 60.96 \\ 
RaCNN-K5-mixup & 75.6 & 46.37 & 37.9 & 28.11 & 74.89 & 74.89 & 48.12 & 66.96 & 63.84 & 63.55 \\
\midrule
RaCNN-K10 & 79.52 & 52.95 & 43.9 & \textbf{33.77} & 79.12 & 79 & \textbf{55.27} & 72.89 & 71.81 & 71.14 \\
RaCNN-K10-mixup & 80.80 & \textbf{53} & \textbf{44.01} & 33.47 &\textbf{ 79.87} & \textbf{79.72} & 54.36 & \textbf{73.63} & \textbf{72.35} & \textbf{71.26} \\

\bottomrule
\end{tabular}
\end{center}
\vskip -0.1in
\end{table*}

\subsection{Attack-Aware Learning}

Another direction has been on modifying a learning algorithm to make a classifier more robust to adversarial examples. As our approach relies on usual backpropagation with stochastic gradient descent, most of the approaches below, as well as above, are readily used together.

\paragraph{Adversarial Training}

Already early on, \citet{goodfellow2014explaining} proposed a procedure of adversarial training, where a classifier is trained on both training examples and adversarial examples generated on-the-fly. \citet{lee2017generative} extended this procedure by introducing a generative adversarial network~\citep[GAN,][]{goodfellow2014generative} that learns to generate adversarial examples while simultaneously training a classifier. These approaches are generally applicable to any system that could be tuned frequently, and could be used to train the proposed model.

\paragraph{Robust Optimization}

Instead of explicitly including adversarial examples during training, there have been attempts to modify a learning algorithm to induce robustness. \citet{cisse2017parseval} proposed parseval training that encourages the Lipschitz constant of each layer of a deep neural network classifier to be less than one. More recently, \citet{sinha2018certifiable} proposed a tractable robust optimization algorithm for training a deep neural net classifier to be more robust to adversarial examples. This robust optimization algorithm ensures that the classifier well-behaves in the neighborhood of each training point. It is highly relevant to the proposed local mixup which also aims at making a classifier well-behave between any pair of neighboring training examples.

\subsection{Retrieval-Augmented Neural Networks}

The proposed approach tightly integrates an off-the-shelf retrieval engine into a deep neural network. This approach of retrieval-augmented deep learning has recently been proposed in various tasks. \citet{gu2017search} use a text-based retrieval engine to efficiently retrieve relevant training translation pairs and let their non-parametric neural machine translation system seamlessly fuse an input sentence and the retrieved pairs for better translation. \citet{wang2017k} proposed a similar approach to text classification, and \citet{guu2017generating} to language modeling. More recently, \citet{sprechmann2018memorybased} applied this retrieval-based mechanism for online learning, similarly to the earlier work by \citet{li2016one} in the context of machine translation.

\begin{figure}[h!]
\small
\minipage{0.48\columnwidth}
\centering
  \includegraphics[width=\linewidth]{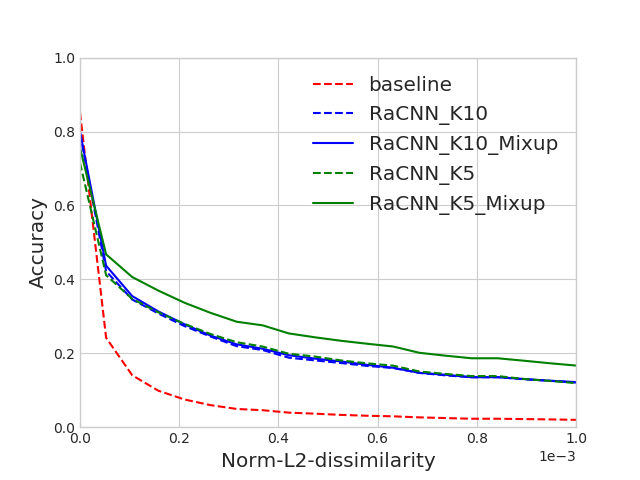}
  
  FGSM
\endminipage\hfill
\minipage{0.48\columnwidth}
\centering
\includegraphics[width=\linewidth]{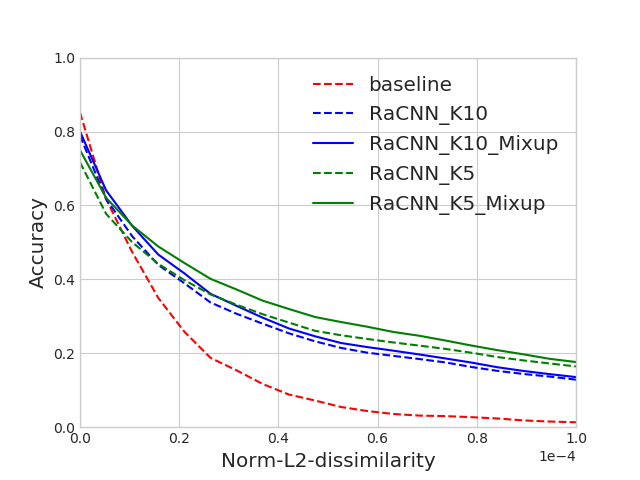}
iFGSM
\endminipage

\minipage{0.48\columnwidth}%
\centering
\includegraphics[width=\linewidth]{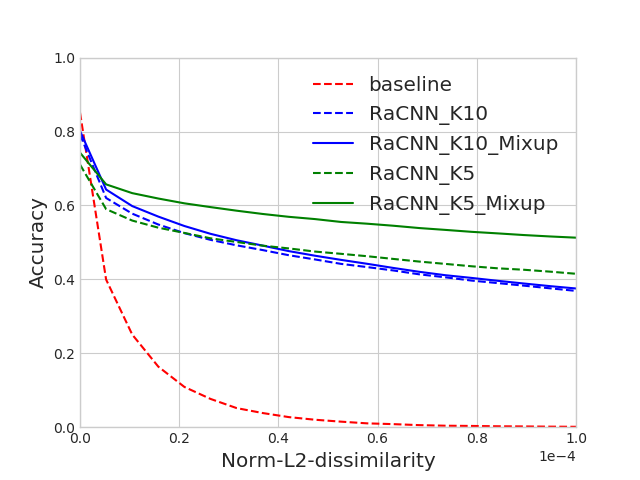}
DeepFool
\endminipage
\hfill
\minipage{0.48\columnwidth}%
\centering
\includegraphics[width=\linewidth]{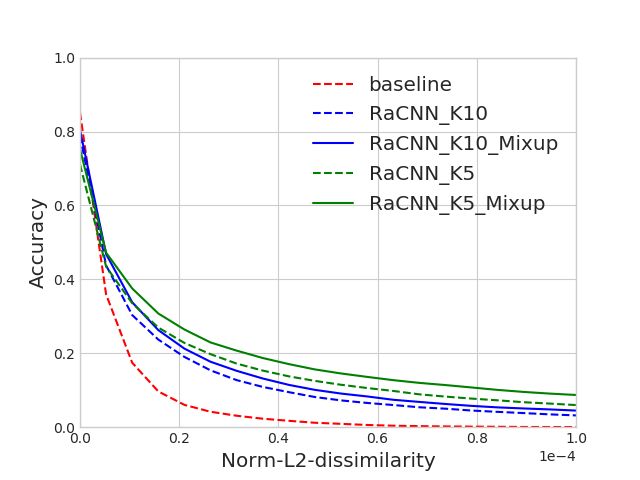}
L-BFGS
\endminipage

\minipage{0.48\columnwidth}%
\centering
\includegraphics[width=\linewidth]{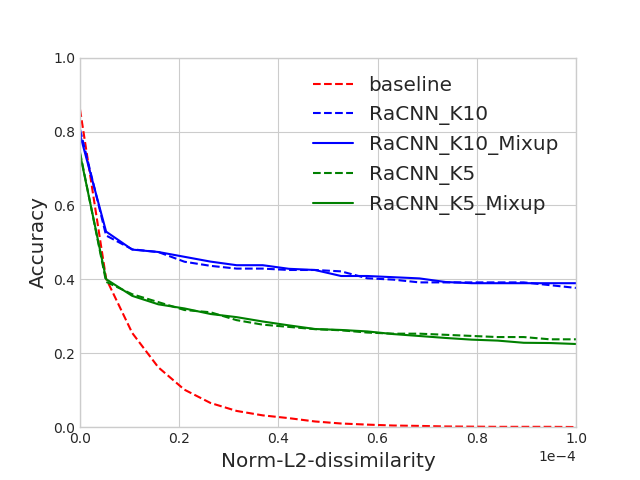}
Boundary
\endminipage\hfill
\minipage{0.48\columnwidth}%
\caption{The CIFAR-10 classifiers' robustness to the adversarial attacks in the Scenario 1 (Direct Attack). The x-axis indicates the strength of attack in terms of the normalized $L_2$ distance. The y-axis corresponds to the accuracy.
\label{fig:cifar10}}
\endminipage

\vspace{-4mm}
\end{figure}

\begin{table*}[h!]
\caption{The SVHN classifiers' robustness to the adversarial attacks in the Scenario 2 (Retrieval Attack)
\label{tab:svhn}}
\begin{center}
\small
\begin{tabular}{c|c|ccc|ccc|ccc} 
\toprule
 & Clean & & FGSM & & &  iFGSM & & & DeepFool \\
\midrule
$\overline{L_2}$ & 0 & 2e-04 & 4e-04 & 8e-04 & 2e-05 & 8e-05 & 2e-04 & 2e-05 & 8e-05  & 2e-04  \\ 
\midrule
Baseline & \textbf{95.48} & 42.09 & 30.95 & 21.61 & 70.41 & 35.53 & 11.17 & 51.10 & 16.00 & 4.28 \\ 
\midrule
RaCNN-K5 & 90.78 & 64.87 & 53.31 & 39.44 & 90.73 & 75.80 & 63.41 & 84.62 & 81.30 & 80.55 \\ 
RaCNN-K5-mixup & 91.64 & 68.31 & 57.20 & \textbf{43.73} & 91.55 & 77.74 & \textbf{65.75} & 86.18& 83.20 & 82.43 \\

\midrule
RaCNN-K10 & 92.19 & 64.94 & 52.24 & 37.73& 92.10 & 76.41 & 62.70 & 86.18 & 84.25 & 82.21 \\
RaCNN-K10-mixup & 92.49 & \textbf{68.72} & \textbf{57.30} & 43.49  & \textbf{92.45} & \textbf{78.26} & 65.50 & \textbf{87.33} & \textbf{84.73} & \textbf{84.10} \\
\bottomrule
\end{tabular}
\end{center}
\vspace{-4mm}
\end{table*}

\section{Experiments}

\subsection{Settings}

\paragraph{Datasets} 

We test the proposed approach (RaCNN) on three datasets of different scales. CIFAR-10 has 50k training and 10k test examples, with 10 classes. SVHN has 73k training and 26k test examples, with 10 classes. ImageNet has 1.3M training and 50k validation examples with 1,000 classes. For CIFAR-10 and ImageNet, we use the original training set as a candidate set, i.e., $D'=D$, while we use the extra set of 531k examples as a candidate set in the case of SVHN. The overall training process involves data augmentation on $D$ but not $D'$.

\paragraph{Pretrained Feature Extractor $\phi'$}

We train a deep convolutional network for each dataset, remove the final fully-connected layers and use the remaining stack as a feature extractor $\phi$ for retrieval. This feature extractor is fixed when used in the proposed RaCNN.

\paragraph{RaCNN: Feature Extractor $\phi$ and Classifier $g$}

We use the same convolutional network from above for the RaCNN as well (separated into $\phi$ and $g$ by the final average pooling) for each dataset. For CIFAR-10 and SVHN, we train $\phi$ and $g$ from scratch. For ImageNet, on the other hand, we fix $\phi=\phi'$ and train $g$ from the pretrained ResNet-18 above. The latter was done, as we observed it greatly reduced training time in the preliminary experiments. 

\paragraph{Training}

We use Adam~\citep{kingma2014adam} as an optimizer. We investigate the influence of the newly introduced components--retrieval and local mixup-- by varying $K\in \left\{ 5, 10\right\}$ and $N_{\text{MU}} \in \left\{0\text{ (no mixup)}, 5\right\}$.

\paragraph{Evaluation}

In addition to the accuracy on the clean test set, we look at the accuracy per the amount of perturbation used to create adversarial examples. We use the default \texttt{MeanSquaredDistance} from the Foolbox library; this amount is computed as a normalized $L_2$ distance between the original example $x$ and its perturbed version $\tilde{x}$:
\begin{align*}
\overline{L_2}(x,\tilde{x}) = \frac{\|x - \tilde{x}\|_2^2}{\text{dim}(x) * \big(\max(x) - \min(x)\big)^2 }.
\end{align*}
We further notice that our attacks are generally performed with clipping the outbounded pixel values at each step.

\subsection{CIFAR-10}

\paragraph{Model}

In the CIFAR-10 experiments, our model contains 6 convolutional layers followed by 2 fully-connected layers. Every layer is operated with batch normalization \citep{ioffe2015batch} and ReLU after. 
More details can be found in Appendix \ref{app:model}.

\paragraph{Scenario 1 (Direct Attack)}
We present in Fig.~\ref{fig:cifar10} the effect of adversarial attacks with varying strengths (measured in the normalized $L_2$ distance) on both the vanilla convolutional network (Baseline) and the proposed RaCNN's with various settings. Across all five adversarial attacks, it is clear that the proposed RaCNN is more robust to adversarial examples than the vanilla classifier is. The proposed local mixup improves the robustness further, especially when the number of retrieved examples is small, i.e., $K=5$. We conjecture that this is due to the quadratically increasing number of pairs, i.e., $\frac{K(K-1)}{2}$, for which local mixup must take care of, with respect to $K$. 

\paragraph{Scenario 2 (Retrieval Attack)}

In Table~\ref{tab:cifar10}, we present the accuracies of both the baseline and RaCNN's with varying strengths of white-box attacks, when the feature extractor $\phi'$ for the retrieval engine is attacked. We observe that it is indeed possible to fool the proposed RaCNN by attacking the retrieval process. Comparing Fig.~\ref{fig:cifar10} and Table~\ref{tab:cifar10}, we however notice that the performance degradation is much less severe in this second scenario.

\begin{figure}[t]
\small

\minipage{0.48\columnwidth}
\centering
  \includegraphics[width=\linewidth]{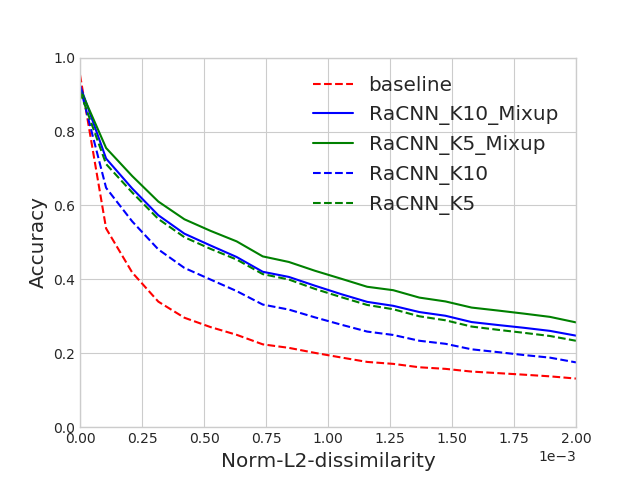}
  FGSM
\endminipage\hfill
\minipage{0.48\columnwidth}
\centering
  \includegraphics[width=\linewidth]{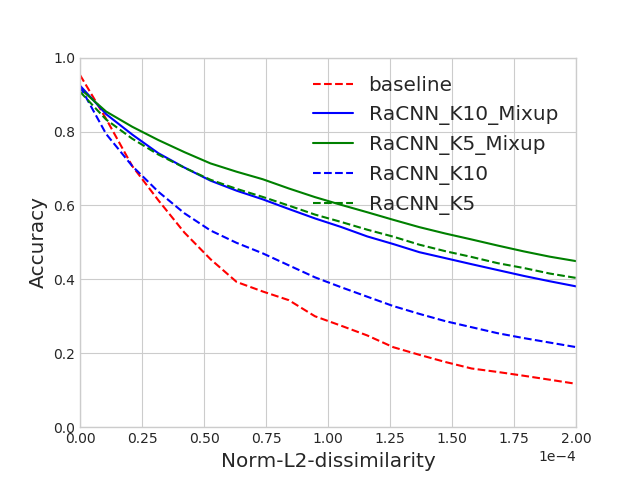}
  iFGSM
\endminipage

\minipage{0.48\columnwidth}%
\centering
  \includegraphics[width=\linewidth]{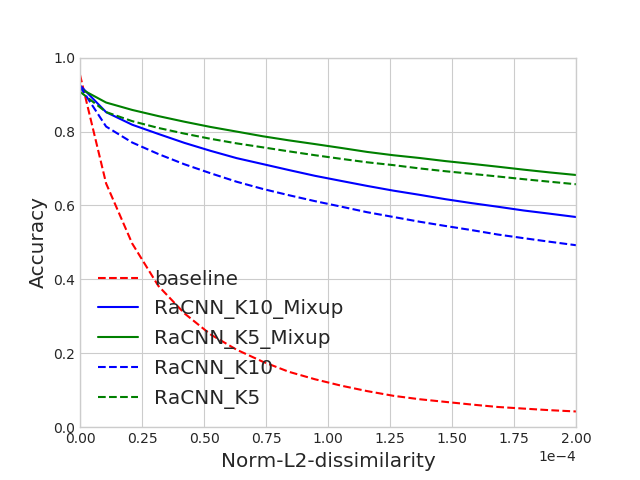}
  DeepFool
\endminipage\hfill
\minipage{0.48\columnwidth}%
\centering
  \includegraphics[width=\linewidth]{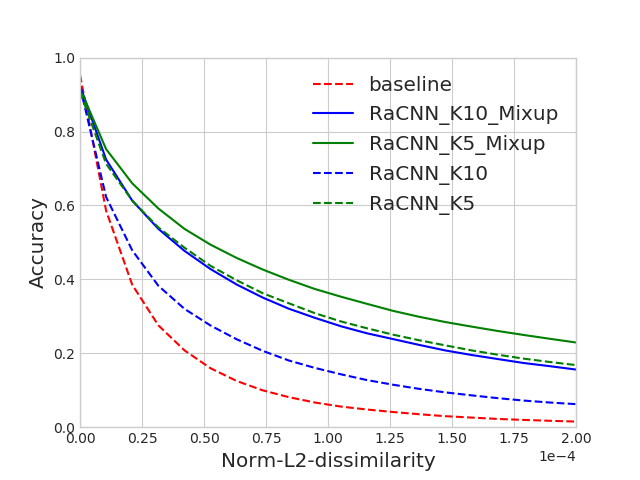}
  L-BFGS
\endminipage

\minipage{0.48\columnwidth}%
\centering
  \includegraphics[width=\linewidth]{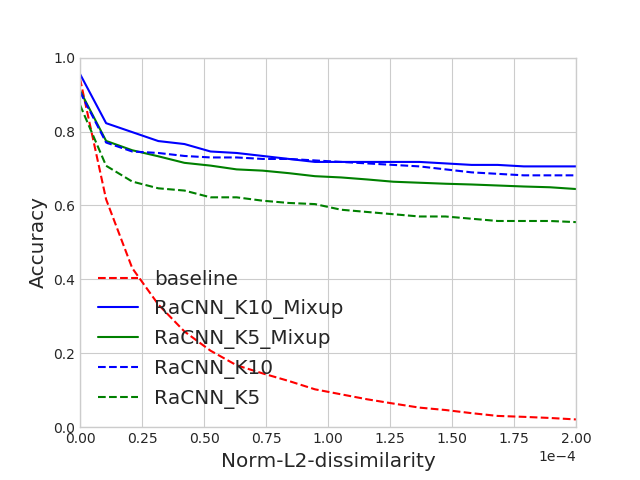}
  Boundary
\endminipage\hfill
\minipage{0.48\columnwidth}%
  \caption{The SVHN classifiers' robustness to the adversarial attacks in the Scenario 1 (Direct Attack). The x-axis indicates the strength of attack in terms of the normalized $L_2$ distance. The y-axis corresponds to the accuracy.
\label{fig:svhn} } 
\endminipage 
\end{figure}

\begin{table*}[h!]
\caption{The ImageNet classifiers' robustness to the adversarial attacks in the Scenario 2 (Retrieval Attack). 
\label{tab:imagenet}}
\begin{center}
\small
\begin{tabular}{c|c|ccc|ccc|ccc} 
\toprule
 & Clean & & FGSM & & &  iFGSM & & & DeepFool \\
\midrule
$\overline{L_2}$ & 0 & 1e-04 & 2e-04 & 4e-04 & 1e-05 & 2e-05 & 4e-05 & 1e-05 & 2e-05  & 4e-05  \\ 
\midrule
Baseline & \textbf{88.98} & 15 & 13.12 & 11.65 & 9.59 & 3.57 & 1.82 & 0.29 & 0.17 & 0.16 \\ 
\midrule
RaCNN-K10-mixup & 77.68 & \textbf{20.17} & \textbf{17.40} & \textbf{14.70}  & \textbf{77.28} & \textbf{64.97} & \textbf{17.67} & \textbf{35.74} & \textbf{35.72} & \textbf{35.71} \\
\bottomrule
\end{tabular}
\end{center}
\vskip -0.1in
\end{table*}

\subsection{SVHN}

\paragraph{Model} 
We use the same architecture and hyper-parameter setting as in the CIFAR-10 experiments.

\paragraph{Scenario 1 (Direct Attack)}

On SVHN, we observe a similar trend from CIFAR-10. The proposed RaCNN is more robust against all the adversarial attacks compared to the vanilla convolutional network. Similarly to CIFAR-10, the proposed approach is most robust to DeepFool and Boundary, while it is most susceptible to L-BFGS. We however notice that the impact of local mixup is larger with SVHN than was with CIFAR-10. 

Another noticeable difference is the impact of the number of retrieved examples on the classification accuracy. In the case of CIFAR-10, the accuracies on the clean test examples (the first column in Table~\ref{tab:cifar10}) between using 5 and 10 retrieved examples differ significantly, while it is much less so with SVHN (the first column in Table~\ref{tab:svhn}.) We conjecture that this is due to a lower level of variation in input examples in SVHN, which are pictures of house numbers taken from streets, compared to those in CIFAR-10, which are pictures of general objects. 

\paragraph{Scenario 2 (Retrieval Attack)}

We observe a similar trend between CIFAR-10 and SVHN, when the feature extractor $\phi'$ for retrieval was attacked, as shown in Tables~\ref{tab:cifar10}--\ref{tab:svhn}.

\begin{figure}[t]
\small

\minipage{0.48\columnwidth}
\centering
  \includegraphics[width=\linewidth]{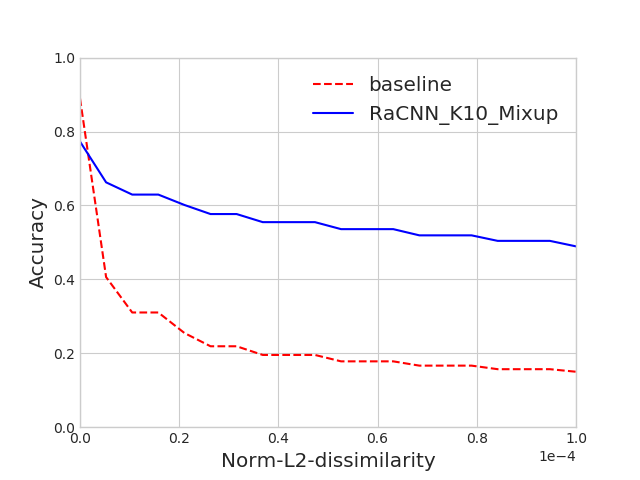}
  FGSM
\endminipage\hfill
\minipage{0.48\columnwidth}
\centering
  \includegraphics[width=\linewidth]{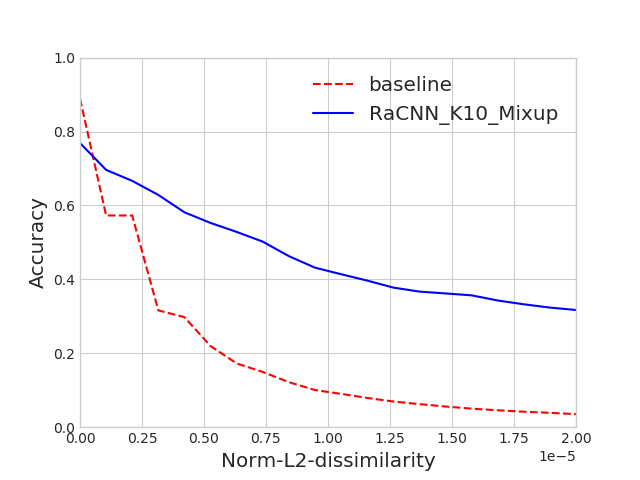}
  iFGSM
\endminipage

\minipage{0.48\columnwidth}%
\centering
  \includegraphics[width=\linewidth]{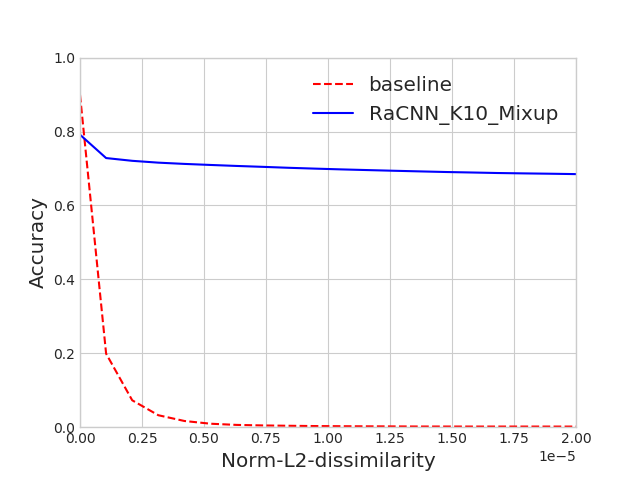}
  DeepFool
\endminipage\hfill
\minipage{0.48\columnwidth}%
\centering
  \includegraphics[width=\linewidth]{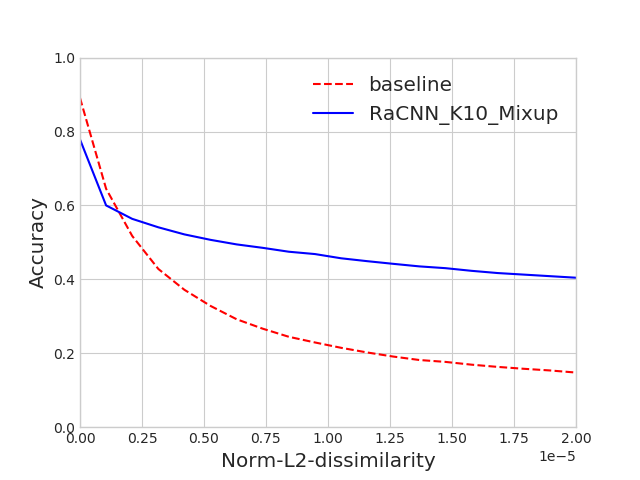}
  L-BFGS
\endminipage

\minipage{0.48\columnwidth}%
\centering
  \includegraphics[width=\linewidth]{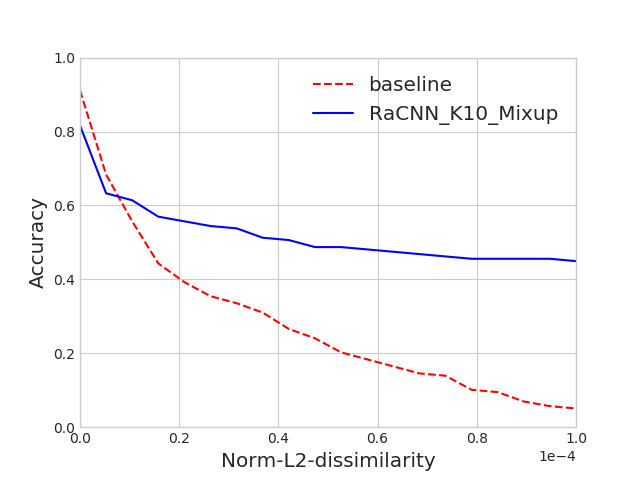}
  Boundary
\endminipage\hfill
\minipage{0.48\columnwidth}%
  \caption{The ImageNet classifiers' robustness to the adversarial attacks in the Scenario 1 (Direct Attack). The x-axis indicates the strength of attack in terms of the normalized $L_2$ distance. The y-axis corresponds to the accuracy. The adversary utilizes top-5 accuracies for attacks.
\label{fig:imagenet} }
\endminipage
\end{figure}

\subsection{ImageNet}

\paragraph{Model}

We use ResNet-18~\citep{he2016deep}. We pretrain it as a standalone classifier on ImageNet and use the feature extractor part $\phi'$ for retrieval. We use the same feature extractor $\phi=\phi'$ for the RaCNN without updating it. The classifier $g$ is initialized with $g'$ and tuned during training. In the case of ImageNet, we only try $K=10$ retrieved examples with local mixup. Due to the high computational cost of the L-BFGS and Boundary attacks, we evaluate both the vanilla classifier and RaCNN against these two attacks on 200 images drawn uniformly at random from the validation set. We use Accuracy@5 which is a standard metric with ImageNet.

\paragraph{Scenario 1 (Direct Attack)}

A general trend with ImageNet is similar to that with either CIFAR-10 or SVHN, as can be seen in Fig.~\ref{fig:imagenet}. The proposed RaCNN is more robust to adversarial attacks. We however do observe some differences. First, iFGSM is better at compromising both the baseline and RaCNN than L-BFGS is, in this case. Second, DeepFool is much more successful at fooling the baseline convolutional network on ImageNet than on the other two datasets, but is much less so at fooling the proposed RaCNN. 


\paragraph{Scenario 2 (Retrieval Attack)}

Unlike CIFAR-10 and SVHN, we have observed that the retrieval attack is sometimes more effective than the direct attack in the case of ImageNet. For instance, FGSM can compromise the retrieval feature extractor $\phi'$ to decrease the accuracy from 77.68 down to 0.20 at $\overline{L_2}=10^{-4}$. We observed a similar behavior with DeepFool, but not with iFGSM. 

\begin{figure*}[h!]
\small
\centering

Clean

\minipage{0.09\textwidth}
  \includegraphics[width=\linewidth]{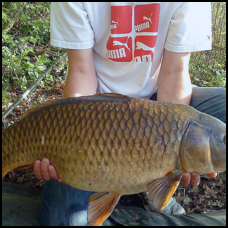}
\endminipage\hfill
\minipage{0.9\textwidth}
  \includegraphics[width=\linewidth]{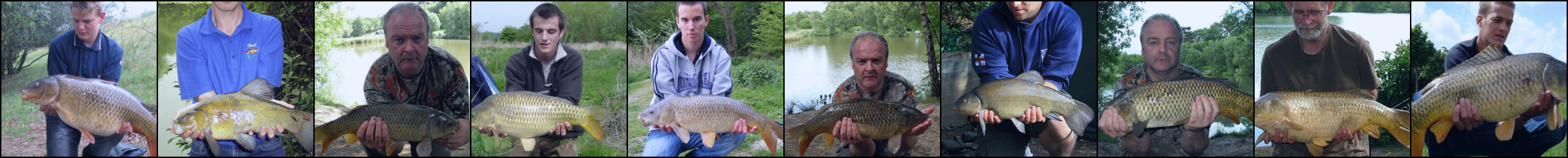}
\endminipage
\vspace{1mm}

iFGSM (Scenario 1 -- Direct Attack) with $\overline{L_2}=2 \times 10^{-5}$

\minipage{0.09\textwidth}
  \includegraphics[width=\linewidth]{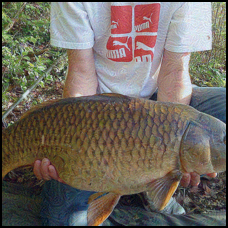}
\endminipage\hfill
\minipage{0.9\textwidth}
  \includegraphics[width=\linewidth]{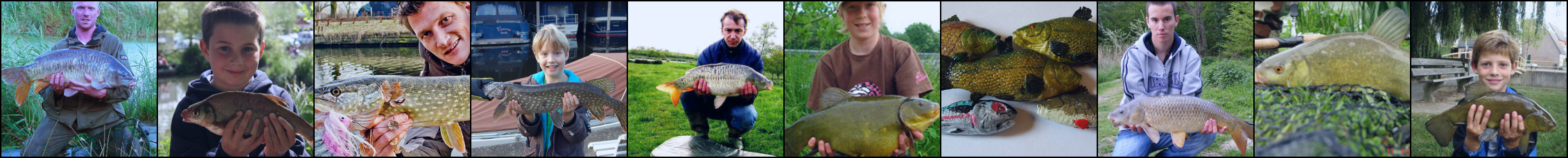}
\endminipage 

iFGSM (Scenario 2 -- Retrieval Attack) with $\overline{L_2}=2 \times 10^{-5}$

\minipage{0.09\textwidth}
  \includegraphics[width=\linewidth]{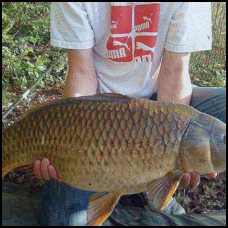}
\endminipage\hfill
\minipage{0.9\textwidth}
  \includegraphics[width=\linewidth]{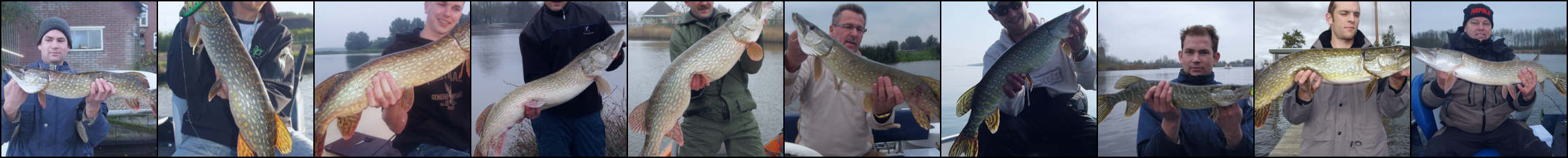}
\endminipage 
\vspace{1mm}

iFGSM (Scenario 1 -- Direct Attack) with $\overline{L_2}=4 \times 10^{-5}$

\minipage{0.09\textwidth}
  \includegraphics[width=\linewidth]{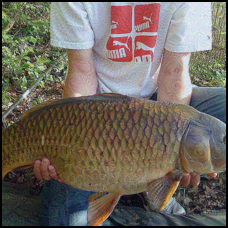}
\endminipage\hfill
\minipage{0.9\textwidth}
  \includegraphics[width=\linewidth]{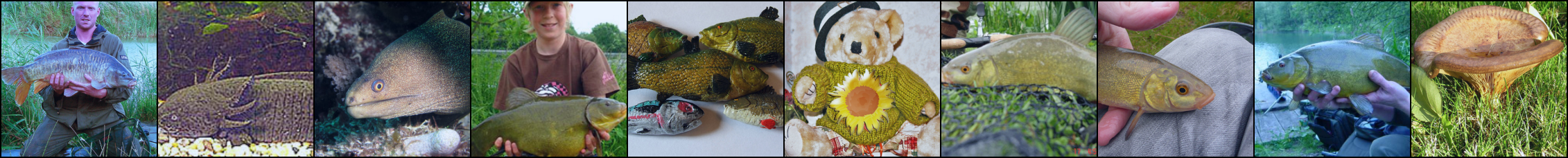}
\endminipage 

iFGSM (Scenario 2 -- Retrieval Attack) with $\overline{L_2}=4 \times 10^{-5}$

\minipage{0.09\textwidth}
  \includegraphics[width=\linewidth]{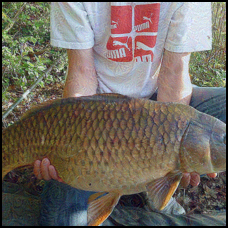}
\endminipage\hfill
\minipage{0.9\textwidth}
  \includegraphics[width=\linewidth]{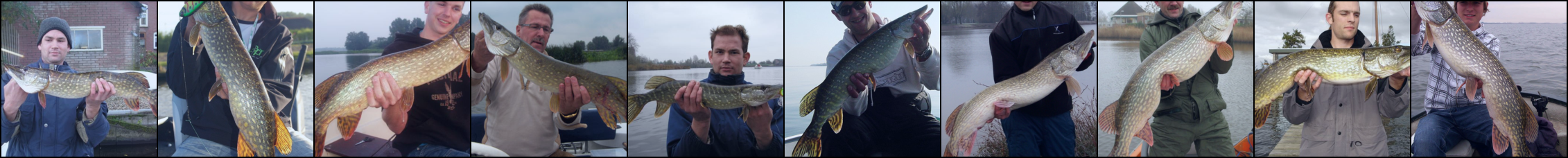}
\endminipage 

\caption{On the left-most column shows the query image, and the next ten images have been retrieved by $F$. We show the retrieval results using the original image and the adversarial images one row at a time. 
With the amount of injected noise high enough to fool any vanilla convolutional network, the behavior of the retrieval engine changes however largely maintains the semantics of the query image. That is, most of the retrieved images contain fish, although specific species may change.
\label{fig:example} }
\vspace{-3mm}
\end{figure*}

\subsection{Discussion}

In summary, we have observed that the proposed RaCNN, when trained with the local mixup, is more robust to adversarial attacks, at least those five considered in the experiments, than the vanilla convolutional network. More specifically, the RaCNN was most robust to the black-box, decision-based attach~\citep{brendel2017decision}, while it was more easily compromised by white-box attacks, especially by the L-BFGS attack~\citep{tabacof2016exploring} which relies on a strong, quasi-Newton optimizer. This suggests that the RaCNN could be an attractive alternative to the vanilla convolutional network when deployed, for instance, in a cloud-based environment. 

In Fig.~\ref{fig:example}, we show retrieval results given a query image from ImageNet. Although adversarial attack did indeed alter the retrieval engine's behavior, we see that the semantics of the original query image could still be maintained in those sets of retrieved images, suggesting two insights. First, the robustness of the RaCNN is largely due to the robustness of the retrieval engine to small perturbation in the input. Even when the retrieval quality degrades, we observe that a majority of retrieved examples are of the same, or a similar, class. Second, we could further improve the robustness by designing the feature extractor $\phi'$ for the retrieval engine more carefully. For instance, an identity function $\phi'(x)=x$ would correspond to retrieval based on the raw pixels, which would make the retrieval engine extremely robust to any adversarial attack imperceptible to humans. This may however results in a lower accuracy on clean examples, which is a trade-off that needs to be determined per task.

As have been observed with the existing input transformation based defense strategies, the robustness of the proposed RaCNN comes at the expense of the generalization performance on clean input examples. We have observed however that this degradation could be controlled at the expense of computational overhead by varying the number of retrieved examples per input. This controllability could be an important feature when deploying such a model in production.

\section{Conclusion}

In this paper, we proposed a novel retrieval-augmented convolutional network classifier (RaCNN) that integrates an off-the-shelf retrieval engine to counter adversarial attacks. The RaCNN was designed to tackle both off- and on-manifold adversarial examples, and to do so, we use a retrieval engine to locally characterize the data manifold as a feature-space convex hull and the attention mechanism to project the input onto this convex hull. The entire model, composed of the retrieval engine and a deep convolutional network, is trained jointly, and we introduced the local mixup learning strategy to encourage the classifier to behave linearly on the feature-space convex hull. 

We have evaluated the proposed approach on three standard object recognition benchmarks--CIFAR-10, SVHN and ImageNet-- against four white-box adversarial attacks and one black-box, decision-based attack. The experiments have revealed that the proposed approach is indeed more robust than the vanilla convolutional network in all the cases. The RaCNN was found to be especially robust to the black-box, decision-based attack, suggesting its potential for the cloud-based deployment scenario. 

The proposed approach consists of three major components; (1) local characterization of data manifold, (2) data manifold projection and (3) regularized learning on the manifold. There is a large room for improvement in each of these components. For instance, a feature-space convex hull may be replaced with a more sophisticated kernel estimator. Projection onto the convex hull could be done better, and a learning algorithm better than local mixup could further improve the robustness against on-manifold adversarial examples. We leave these possibilities as future work.



\bibliography{icml}
\bibliographystyle{icml2018}

\clearpage

\section{Appendix: Model details}
\label{app:model}

Our CIFAR-10 and SVHN model spec is listed in the following table.
\begin{table}[h!]
\begin{center}
\footnotesize
\begin{tabular}{c|p{0.50\columnwidth}|c}
\toprule
\textbf{Stage} & \textbf{Architecture} & \textbf{Size} \\
\midrule
Feature extractor & 96 3x3 convolution & 96 x 30 x 30\\
		$\phi$	  & batch normalization\\
                  & 96 3x3 convolution & 96 x 28 x 28 \\
                  & batch normalization \& ReLU \\
                  & 96 3x3 convolution with stride 2x2 & 96 x 13 x 13 \\
                  & batch normalization \& ReLU \\
                  & 192 3x3 convolution & 192 x 11 x 11 \\
                  & batch normalization \& ReLU \\
                  & 192 3x3 convolution with stride 2x2 &  192 x 4 x 4  \\
                  & batch normalization \\
\midrule
Attention & 	  256 4x4 convolution & 256 \\
 &                Convex-sum (with attention mechanism $U$ or local mixup)  & 256 \\
\midrule
Classification  & fully-connected layer 256 x 64 & 64 \\
				& batch normalization \& ReLU \\
				& fully-connected layer 64 x 10 & 10 \\
\bottomrule
\end{tabular}
\end{center}
\end{table}

The pretrained retrieval index building $\phi'$ network is listed as follow:
\begin{table}[h!]
\begin{center}
\footnotesize
\begin{tabular}{c|p{0.50\columnwidth}|c}
\toprule
\textbf{Stage} & \textbf{Architecture} & \textbf{Size} \\
\midrule
Feature extractor & 96 3x3 convolution & 96 x 30 x 30\\
		$\phi'$	  & batch normalization \& ReLU\\
                  & 96 3x3 convolution & 96 x 28 x 28 \\
                  & batch normalization \& ReLU \\
                  & 96 3x3 convolution with stride 2x2 & 96 x 13 x 13 \\
                  & batch normalization \& ReLU \\
                  & 192 3x3 convolution & 192 x 11 x 11 \\
                  & batch normalization \& ReLU \\
                  & 192 3x3 convolution with stride 2x2 &  192 x 4 x 4  \\
                  & batch normalization \\
\midrule
Classification  & fully-connected layer 3072 x 512 & 512 \\
\textbf{Used only} & batch normalization \& ReLU \\
\textbf{ in Scenario 2}	& fully-connected layer 512 x 128 & 128 \\
               & batch normalization \& ReLU & \\
          		& fully-connected layer 128 x 10 & 10 \\
\bottomrule
\end{tabular}
\end{center}
\end{table}

\end{document}